\documentclass[conference]{IEEEtran}
\IEEEoverridecommandlockouts
\usepackage{cite}
\usepackage{amsmath,amssymb,amsfonts}
\usepackage{algorithmic}
\usepackage{graphicx}
\usepackage{textcomp}
\usepackage{xcolor}
\usepackage{booktabs}
\usepackage{tikz}
\usetikzlibrary{positioning, fit, backgrounds, arrows.meta, calc}

\pgfdeclarelayer{zoneback}
\pgfsetlayers{zoneback,background,main}

\def\BibTeX{{\rm B\kern-.05em{\sc i\kern-.025em b}\kern-.08em
    T\kern-.1667em\lower.7ex\hbox{E}\kern-.125emX}}
\begin{document}

\title{A Deployed Hybrid Vehicle-in-the-Loop Platform for Validating Cooperative Perception
\thanks{This work was supported by the SYNERGIES project, funded by the European Union's Horizon Europe research and innovation programme under Grant Agreement No. 101146542. Views and opinions expressed are however those of the author(s) only and do not necessarily reflect those of the European Union or CINEA. Neither the European Union nor the granting authority can be held responsible for them.}
}

\author{
\IEEEauthorblockN{
Anastasia Bolovinou, Giorgos Hadjipavlis, Markos Antonopoulos, Panagiotis Tachtalis, Konstantinos Petousakis,\\Konstantinos Lazaridis,Alexandros Siskos, Bill Roungas, Angelos Amditis}
\IEEEauthorblockA{\textit{Institute of Communication and Computer Systems, Athens, Greece}\\
\{anastasia.bolovinou, giorgos.hadjipavlis, panagiotis.tachtalis, markos.antonopoulos, konstantinos.petousa, k.lazaridis,\\alexandros.siskos, vroungas, a.amditis\}@iccs.gr}
}

\maketitle

\begin{abstract}
European safety regulation now permits a large share of automated-driving homologation evidence to be produced virtually, provided a validated physical-virtual facility generates it. We present a deployed hybrid Vehicle-in-the-Loop (ViL) platform that couples a real instrumented vehicle with a CARLA-based digital twin (DT) through a V2X message pipeline, and we report its first integrated operation on a public-road-representative test track. A real vehicle streams ETSI-compliant CAM/CPM messages into the DT, where a GPU-accelerated Cooperative Perception (CP) module fuses them into a probabilistic occupancy grid during scenario runtime. We demonstrate the platform on a multi-vehicle double T-intersection scenario, characterise the CP workload across nominal, rain and night conditions and five localization-noise levels, and discuss the platform's current architectural limits and the engineering targets they define. The results show that CP substantially widens field-of-view (FoV) coverage and improves occupied-cell recall, and that beyond a moderate localization-noise threshold, positioning uncertainty, and not weather, becomes the dominant error source. We outline the platform's trajectory toward a Mediterranean operational design domain (ODD) testing service.
\end{abstract}


\section{Introduction}

The EU General Safety Regulation \cite{gsr2} has made advanced driver-assistance systems mandatory across all new vehicle categories, while recent UN regulations for automated systems (UN R152, R157 and R171) \cite{unr152,unr157,unr171} now accept virtual testing as valid homologation evidence, on the condition that the virtual results are validated against physical tests at the same facility. A testing facility that couples a real vehicle to a credible simulation, rather than a pure simulator, is therefore precisely what the regulatory framework requires.

Hybrid testing is also a practical necessity. Driving scenarios, including CP scenarios, may require more connected agents than an organization physically owns, and the mix of real and virtual participants varies from experiment to experiment with the available hardware, like the number of instrumented vehicles, road side units, extra traffic agents (connected or not connected) etc. By substituting virtual agents for unavailable real ones, a hybrid platform can exercise multi-agent scenarios that would otherwise be infeasible to stage physically, at a virtual/real ratio chosen per experiment.
 
This paper describes such a facility as deployed infrastructure. Our contribution is threefold:

\begin{itemize}
\item A hybrid ViL platform that couples a real instrumented vehicle to a CARLA digital twin (DT) through an ETSI-compliant Cooperative Awareness Messages (CAM)/Collective Perception Messages (CPM) pipeline \cite{etsicam,etsicpm} extended from the CARLA ROS bridge.
\item A real-world demonstrator on 
our institution's public-road-representative test track, in which three real vehicles (two sharing CPM) are reflected in the DT and consumed by a GPU-accelerated occupancy-grid module.
\item A characterisation of the CP workload across nominal, rain and night conditions, together with a statement of the platform's current limits.
\end{itemize}

We deliberately separate what is demonstrated functionally on the real vehicle from what is quantified in simulation; closing that gap with a quantitative sim-to-real fidelity metric is ongoing work.

\begin{figure*}[t]
  \centering
  \includegraphics[width=1.6\columnwidth]{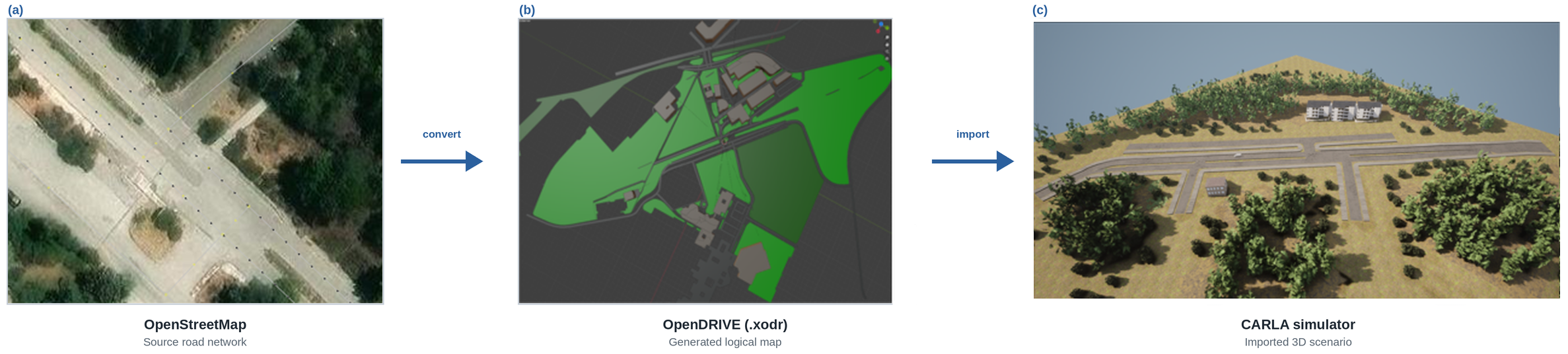}
  \caption{Twin-construction pipeline: (a) the source road network in OpenStreetMap is (b) converted to an OpenDRIVE (.xodr) logical map and (c) imported into CARLA as a drivable 3D scenario.}
  \label{fig:pipeline}
\end{figure*}

\section{Platform Architecture}

\begin{figure}[t]
  \centering
  \resizebox{0.8    \columnwidth}{!}{%
  \begin{tikzpicture}[
    font=\scriptsize, >=Latex,
    node distance=2.5mm and 4mm,
    box/.style={rectangle, rounded corners=2pt, draw, fill=gray!8,
                align=center, inner sep=3pt, minimum height=8mm, text width=21mm},
    outbox/.style={box, fill=gray!22},
    softbox/.style={box, densely dashed, fill=gray!3},
    grp/.style={draw=black!55, dashed, rounded corners=3pt, inner sep=6pt},
    zone/.style={rounded corners=4pt, inner sep=10pt},
    lbl/.style={font=\scriptsize\itshape, text=black!70},
    zlbl/.style={font=\scriptsize\bfseries, text=black!55},
    flow/.style={->, thick}
  ]
    \node[box] (sens) {GNSS/IMU,\\ VLP-16 LiDAR,\\ RGB camera};
    \node[box, right=of sens] (nuc) {NUC edge PC\\ (on-board\\ detection)};
    \node[box, right=of nuc] (modem) {Nfiniity Cube\\ V2X modem};
    \draw[flow] (sens) -- (nuc);
    \draw[flow] (nuc) -- (modem);
    \begin{scope}[on background layer]
      \node[grp, fit=(sens)(nuc)(modem)] (veh) {};
    \end{scope}
    \node[lbl, anchor=south west] at (veh.north west) {Real assets (vehicle, RSUs)};
    \node[box, below=20mm of sens] (mid) {Middleware\\ (extends CARLA\\ ROS bridge)};
    \node[box, right=of mid] (twin) {CARLA\\ digital twin};
    \node[box, right=of twin] (agents) {VaN3Twin agents\\ (ns-3 +\\ OpenCDA)};
    \draw[flow] (mid) -- (twin);
    \draw[flow] (agents) -- (twin);
    \begin{scope}[on background layer]
      \node[grp, fit=(mid)(twin)(agents)] (dt) {};
    \end{scope}
    \node[lbl, anchor=south west] at (dt.north west) {Virtual assets (digital twin)};
    \node[softbox, below=16mm of mid] (cam) {Camera--LiDAR\\ perception\\ (supported)};
    \node[box, right=of cam] (occ) {``Occupancy'' CP\\ (Docker $\cdot$ ROS\,2\\ $\cdot$ JAX/CUDA)};
    \node[outbox, right=of occ] (grid) {nav\_msgs/\\ OccupancyGrid};
    \draw[flow] (occ) -- (grid);
    \begin{scope}[on background layer]
      \node[grp, fit=(cam)(occ)(grid)] (cp) {};
    \end{scope}
    \node[lbl, anchor=south west] at (cp.north west) {ROS-based perception modules};
    \begin{pgfonlayer}{zoneback}
      \node[zone, fill=blue!8, fit=(dt)(cp), yshift=6pt] (cloud) {};
    \end{pgfonlayer}
    \node[zlbl, anchor=south east, xshift=-176pt, yshift=-2pt] at (cloud.north east) {Off-board (edge/cloud)};
    \draw[flow] (veh.south) -- node[above right, lbl, fill=white, inner sep=1.5pt, xshift=2pt, yshift=6pt] {ETSI-compliant CAM/CPM (100\,ms)} (dt.north);
    \draw[flow] (dt.south)  -- node[above right, lbl, fill=blue!8, inner sep=1.5pt, xshift=2pt, yshift=-2pt] {fused CAM/CPM streams} (occ.north);
  \end{tikzpicture}}
  \caption{End-to-end architecture of the hybrid ViL platform.}
  \label{fig:arch}
\end{figure}
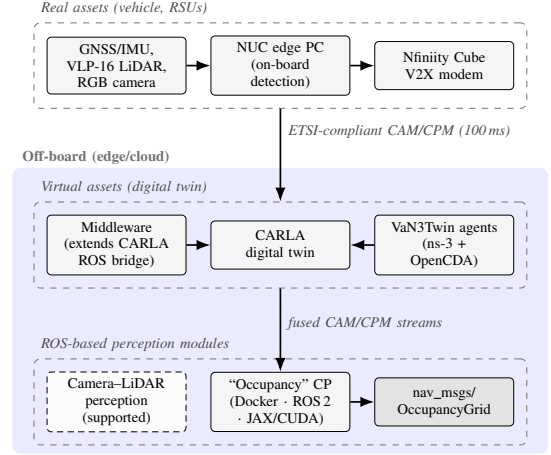

The platform comprises three groups of components (Fig. \ref{fig:arch}): real assets, virtual assets and the ROS-based perception modules that run across both.

\textbf{Real assets.} The core real asset is a fully instrumented vehicle carrying GNSS and an IMU for localization (supporting both standard GPS and differential GPS (D-GPS), the latter providing centimetre-level positioning usable as ground truth), a Velodyne VLP-16 LiDAR and a forward-facing RGB camera for perception, a NUC-class edge PC for on-board object detection, and an Nfiniity Cube EVK OnBoard Unit (OBU) providing C-V2X and ITS-G5 connectivity. The platform further supports additional connected vehicles and roadside units (RSUs) as real assets. Connected assets transmit their kinematic state as CAM and their detections as CPM \cite{etsicam, etsicpm}, both triggered every 100 ms. While the platform is currently tested in a V2X CP configuration, the ViL coupling itself is agnostic to the communication layer; assets and the DT can equally be linked over a direct vehicle-to-cloud connection. Finally, the platform also supports a driver-in-the-loop configuration, in which a human drives the real vehicle while the DT and its virtual agents react in real time, closing the loop through the human as well as the sensors.

\begin{figure}[t]
  \centering
  \includegraphics[width=0.85\columnwidth]{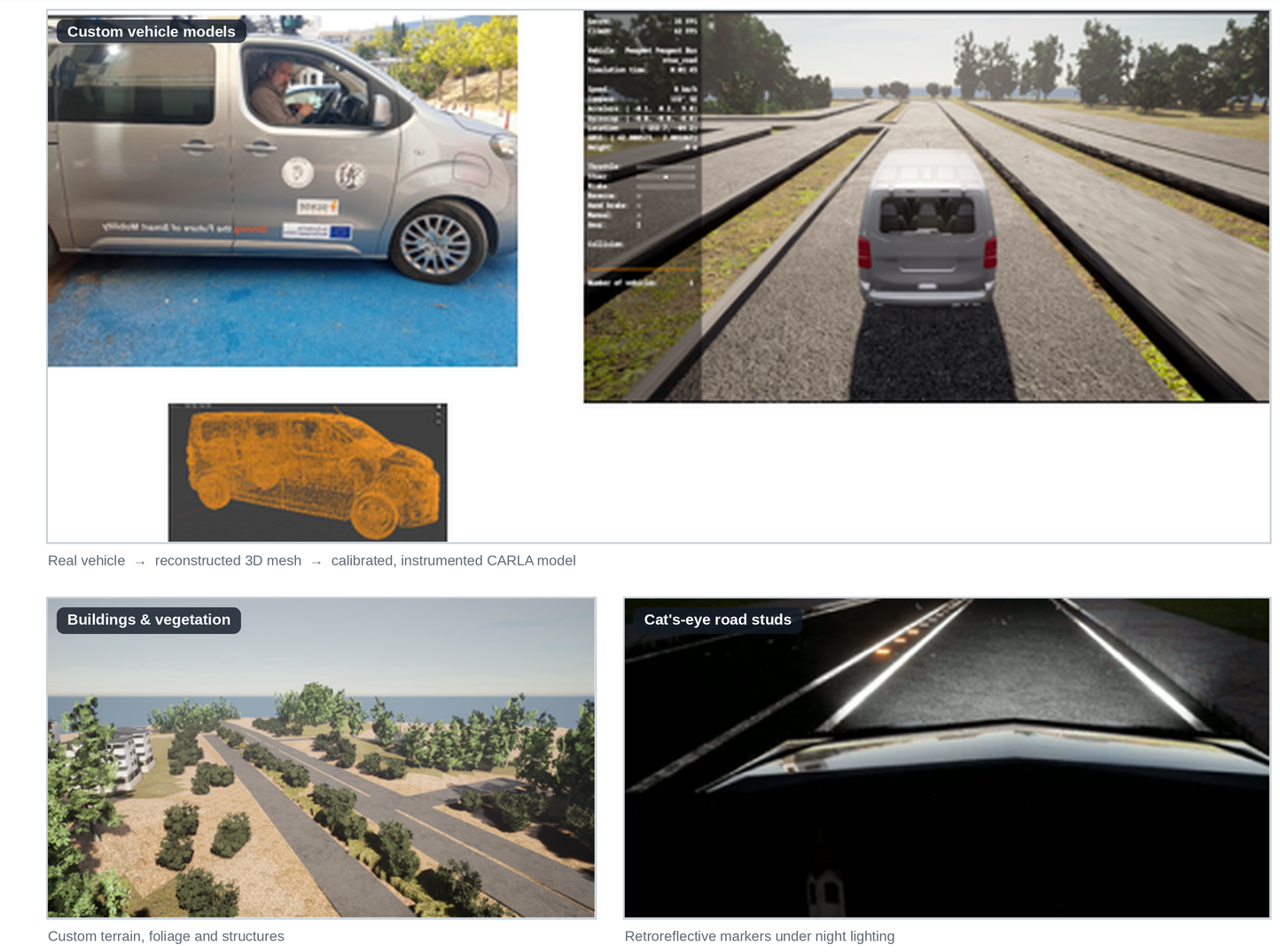}
  \caption{Custom twin assets. Top: the real vehicle is reconstructed as a 3D
  mesh and imported as a calibrated, instrumented CARLA model. Bottom: custom
  terrain, buildings and vegetation (left) and retroreflective cat's-eye road
  studs under night lighting (right).}
  \label{fig:assets}
\end{figure}

\textbf{Virtual assets.} The DT is not merely a drivable map but a complete model of the road environment, its static and dynamic assets, and the simulated agents that populate it, all kept consistent with the real world. A CARLA \cite{carla} instance of this environment ingests the live stream through a custom middleware layer that extends the CARLA ROS bridge, parsing ETSI messages into DT-compatible objects. Virtual agents participate in V2X through the VaN3Twin framework (ns-3 + OpenCDA) \cite{pegurri2026van3twin,xu2021opencda}, so that simulated and real entities both contribute to the shared perception. The site is reconstructed through a repeatable pipeline (Fig. \ref{fig:pipeline}): the road network is extracted from OpenStreetMap, enriched, and converted into an OpenDRIVE (\texttt{.xodr}) logical map encoding lanes, junctions and connectivity, and imported into CARLA as a newly created unreal map (.umap file). The same toolchain reconstructs any OpenStreetMap-covered site as a DT, which is what makes the Mediterranean-ODD scenario library of Section \ref{sec:roadmap} tractable to scale beyond a single location. Validity also depends on the DT reproducing the features that drive sensor and perception behaviour, so the default CARLA asset set is augmented with site- and vehicle-specific models (Fig. \ref{fig:assets}): the real vehicle is reconstructed as a calibrated 3D mesh whose geometry and on-board sensor placement are modelled on the physical platform; the site's buildings, terrain and vegetation are modelled to reproduce real occlusions and sight-lines; condition-specific features such as retroreflective cat's-eye road studs recreate night-time appearance cues. These assets directly underpin the multi-condition behaviour of Section \ref{sec:cp}: the night results, in particular, are only meaningful if the DT renders the visual environment a vehicle would encounter after dark.

\textbf{ROS-based perception modules.} The platform hosts its perception as ROS modules: a single-view camera--LiDAR perception module and the cooperative-perception ``Occupancy'' module (containerised; ROS2; Google JAX with NVIDIA CUDA via XLA JIT), which fuses incoming CAM/CPM into a standardised \texttt{nav\_msgs/msg/OccupancyGrid} -- a probabilistic grid whose cells encode occupancy likelihood (occupied, free, or uncertain), regenerated for each batch of messages. Because both are ROS nodes consuming standardised messages, they deploy unchanged on the real vehicle or against the DT through the CARLA-ROS bridge; this paper evaluates the CP module (Section \ref{sec:cp}).

\section{Real-World Demonstrator}

\begin{figure*}
  \centering
  \includegraphics[width=0.65\linewidth]{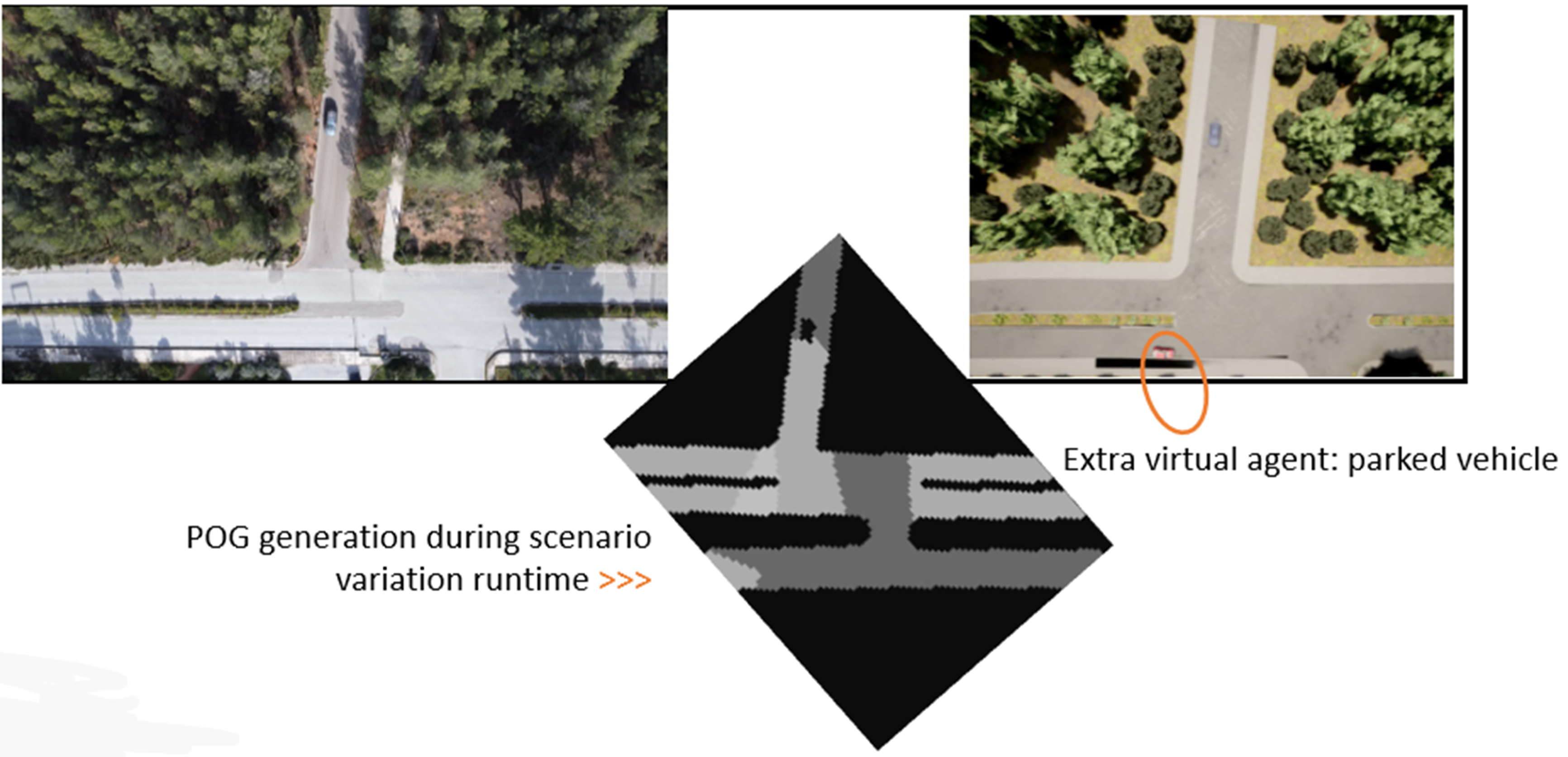}
  \caption{Left: drone footage of the real double T-intersection. Right: CARLA DT reconstruction. Center bottom: occupancy grid produced from the live CAM/CPM stream (with one virtual agent added).}
  \label{fig:demo}
\end{figure*}

We exercised the full chain on 
our institution's test track using a double T-intersection scenario. Three real vehicles operated on the track (two connected and sharing CPM and a third, CAM-only vehicle broadcasting its position but not perception data) while the same scenario was reconstructed in the DT. Fig. \ref{fig:demo} places, side by side, drone footage of the real intersection, its CARLA reconstruction, and the occupancy grid produced from the live messages, including a variation in which a virtual agent augments the real ones.

This demonstrator validates the platform functionally: real CAM/CPM traffic is ingested, reflected in the DT and turned into a coherent occupancy grid end to end. We emphasise its scope honestly. Because the test-track vehicle cannot be synchronised frame-exactly with the simulation, repeated runs of the same initial conditions are required, and the perception metrics in Section \ref{sec:cp} are computed in the DT with controlled localization noise rather than against real-vehicle ground truth. The demonstrator is thus a proof of concept for the hybrid methodology; quantitative sim-to-real fidelity is the subject of follow-up work.

The platform is not a one-off demonstrator but shared infrastructure already exercised across EU projects. In EVENTS \cite{events}, it supports collective perception evaluation and the scaling-up of cooperative scenarios. In SUNRISE \cite{sunrise}, the same hybrid setup is used for hybrid (virtual-physical) testing and the logging of key performance indicators feeding CCAM safety argumentation (video available on request).
\footnote{See details at 6:15 in the following video: https://www.youtube.com/watch?v=OClo941WAso}
The same CARLA-based simulation toolchain, including its driver-in-the-loop configuration, also underpins scenario-data generation in SYNERGIES \cite{synergies}, for the project's federated scenario dataspace.

\section{CP Workload and Multi-Condition Behaviour} \label{sec:cp}

The platform itself produces no metric in isolation; we therefore test it through the CP module it runs. The derived occupancy-grid is evaluated, using the KPIs as a demo for what the platform is capable of, in the DT under three environmental conditions (nominal, rain, night), three cooperative configurations (ego-only, three connected agents, all six agents), and five localization-noise levels (0, 0.5, 1.0, 1.5 and 2.0 meters). All metrics are reported as area-under-curve (AUC) over each scenario.

\begin{figure}
  \centering
  \includegraphics[width=\linewidth]{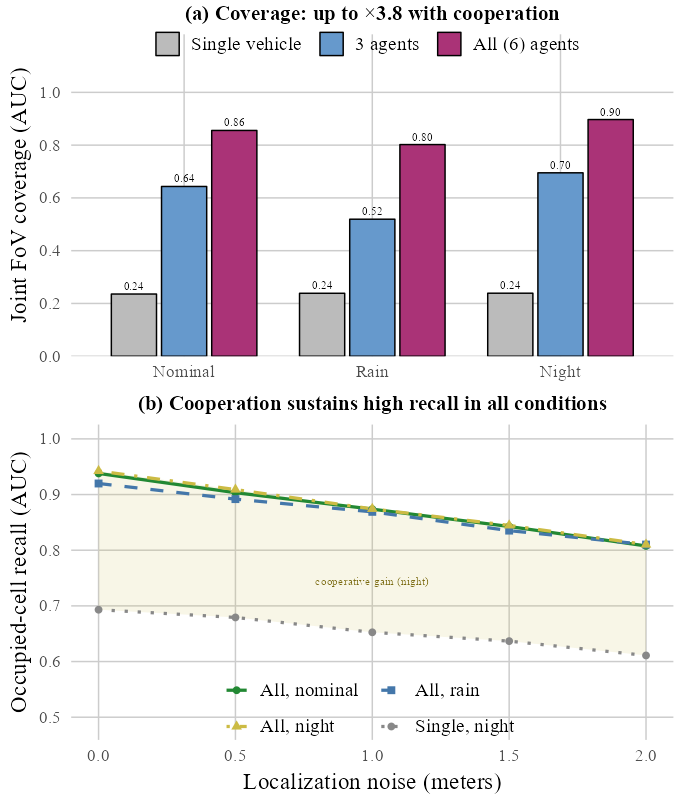}
  \caption{Top (a): Joint FoV coverage of CP across conditions. Bottom (b): Occupied-cell recall across different environmental conditions and localization noise.}
  \label{fig:recall}
\end{figure}

Fig. \ref{fig:recall} summarises the headline KPIs, from which three findings stand out. First, joint FoV coverage rises sharply with cooperation, from roughly one quarter of the area of interest for a single vehicle to the majority of it with all agents (Fig. \ref{fig:recall}a) and, notably, stays stable across the full localization-noise range, indicating that probabilistic fusion preserves spatial awareness even under positioning uncertainty. Second, occupied-cell recall improves with cooperation at low noise but all configurations (except ego-only) converge near the same value at 2 meters (Fig. \ref{fig:recall}b): beyond a moderate noise threshold, localization uncertainty (not the number of agents or the weather) becomes the dominant error source. Third, the cooperative benefit is largest exactly when individual sensing is weakest, i.e., under night conditions, where adding agents recovers occupied-cell recall from a degraded single-vehicle level back toward its nominal value (shaded region in Fig. \ref{fig:recall}b). Occupied-cell precision shows the expected slight decrease as agents are added (overlapping noisy observations), an acceptable trade-off in a safety setting where missed obstacles matter more than conservative ones; free-space estimation remains reliable across all conditions.

\section{Current Limitations and Engineering Targets}

With regard to the platform's current limitations, the coupled pipeline executes on a fixed-interval cycle, where each cycle ingests the live CAM/CPM stream, updates the DT and regenerates the occupancy grid. Still, strict determinism cannot be guaranteed in the current configuration, because transient slowdowns of the simulation loop introduce timing irregularities that can alter the temporal alignment between real-vehicle message arrivals and simulation state updates. This is an architectural property of the CARLA-based co-simulation rather than a hardware limit.

With regard to the engineering targets, the first one is raising the loop rate (the frequency of the complete ingest-update-regenerate cycle) and bounding its jitter (the worst-case deviation of the cycle interval), since bounded timing is a precondition for run-to-run reproducibility in a testing context. We are pursuing this through a co-simulation scheduler that advances all components on a common clock and by isolating the grid generator from the simulator's rendering loop. The second target  is a GPU-versus-CPU benchmark of the occupancy module itself, to quantify the contribution of the JAX/CUDA implementation. The planned sim-to-real fidelity assessment builds directly on this: a D-GPS reference trajectory provides the positioning ground truth against which the DT's synthetic localization noise can be replaced by measured real-vehicle error. Planned extensions broaden the loop further: remote operation of the real vehicle and an immersive driver interface via VR headset for in-the-loop scenario testing.

\section{Toward a Mediterranean-ODD Testing Service}\label{sec:roadmap}

No hybrid ViL facility currently operates in southeastern Europe, to our knowledge. Moreover, existing European facilities are optimised for established homologation procedures rather than hybrid V2X-in-the-loop testing. The platform's planned differentiator is a validated scenario library encoding Mediterranean ODDs (high solar glare, high ambient temperatures, dry-road friction profiles, dense mixed urban traffic) built on the architecture described here. Combined with a progressive accreditation path (ISO 9001, then ISO/IEC 17025), this positions the platform as deployable testing infrastructure for suppliers and developers with the intention to complement existing commercial proving grounds and EuroNCAP test sites, adding hybrid V2X-in-the-loop capability on our institution's own 5G-enabled testbed.

\section{Conclusion}

We presented a deployed hybrid ViL platform, its first integrated real-world demonstrator and a multi-condition characterisation of the CP workload it runs. The platform turns live V2X traffic into a probabilistic occupancy grid end to end, and the evaluation shows that cooperation widens coverage and improves recall while localization uncertainty dominates beyond a moderate threshold. Ongoing work targets quantitative sim-to-real fidelity, a higher and bounded loop rate and the Mediterranean-ODD scenario library.


\bibliographystyle{IEEEtran}
\bibliography{biblio}

@misc{gsr2,
  author       = {{European Parliament and Council of the European Union}},
  title        = {Regulation ({EU}) 2019/2144 on type-approval requirements for motor vehicles ({General Safety Regulation})},
  howpublished = {Official Journal of the European Union},
  year         = {2019}
}

@misc{unr152,
  author       = {{UNECE}},
  title        = {{UN} Regulation No.~152 -- Uniform provisions concerning the approval of motor vehicles with regard to the {Advanced Emergency Braking System (AEBS)} for {M1} and {N1} vehicles},
  howpublished = {United Nations Economic Commission for Europe},
  year         = {2020}
}

@misc{unr157,
  author       = {{UNECE}},
  title        = {{UN} Regulation No.~157 -- Automated Lane Keeping Systems (ALKS) extended to trucks, buses and coaches},
  howpublished = {United Nations Economic Commission for Europe},
  year         = {2022}
}

@misc{unr171,
  author       = {{UNECE}},
  title        = {{UN} Regulation No.~171 -- Uniform provisions concerning the approval of vehicles with regard to {Driver Control Assistance Systems (DCAS)}},
  howpublished = {United Nations Economic Commission for Europe},
  year         = {2024}
}

@manual{etsicam,
  author       = {{ETSI}},
  title        = {Intelligent Transport Systems ({ITS}); Vehicular
                  Communications; Basic Set of Applications; Part~2:
                  Specification of Cooperative Awareness Basic Service},
  organization = {European Telecommunications Standards Institute},
  note         = {{ETSI} EN 302 637-2}
}

@manual{etsicpm,
  author       = {{ETSI}},
  title        = {Intelligent Transport Systems ({ITS}); Vehicular
                  Communications; Basic Set of Applications; Collective
                  Perception Service},
  organization = {European Telecommunications Standards Institute},
  note         = {{ETSI} TS 103 324}
}

@inproceedings{carla,
  author       = {Dosovitskiy, Alexey and Ros, German and Codevilla, Felipe
                  and L{\'o}pez, Antonio and Koltun, Vladlen},
  title        = {{CARLA}: An Open Urban Driving Simulator},
  booktitle    = {Proc. 1st Annual Conf. on Robot Learning (CoRL)},
  pages        = {1--16},
  year         = {2017}
}

@article{pegurri2026van3twin,
  author  = {Pegurri, Roberto and Gasco, Diego and Linsalata, Francesco and
             Rapelli, Marco and Moro, Eugenio and Raviglione, Francesco and
             Casetti, Claudio},
  title   = {{VaN3Twin}: The Multi-Technology {V2X} Digital Twin with
             Ray-Tracing in the Loop},
  journal = {IEEE Transactions on Wireless Communications},
  year    = {2026},
  volume  = {25},
  pages   = {16812--16827},
  doi     = {10.1109/TWC.2026.3531234}
}

@inproceedings{xu2021opencda,
  title={Opencda: an open cooperative driving automation framework integrated with co-simulation},
  author={Xu, Runsheng and Guo, Yi and Han, Xu and Xia, Xin and Xiang, Hao and Ma, Jiaqi},
  booktitle={2021 IEEE International Intelligent Transportation Systems Conference (ITSC)},
  pages={1155--1162},
  year={2021},
  organization={IEEE}
}

@misc{events,
  author       = {{EVENTS Project}},
  title        = {{EVENTS}: reliabl{E} in-{V}ehicle p{E}rception and decisio{N}-making in complex environmen{T}al condition{S}},
  howpublished = {Horizon Europe project, Grant Agreement No.~101069614},
  note         = {\url{https://events-project.eu}}
}

@misc{sunrise,
  author       = {{SUNRISE Project}},
  title        = {{SUNRISE}: Safety Assurance Framework for Connected and
                  Automated Mobility Systems},
  howpublished = {Horizon Europe project, Grant Agreement No.~101069573},
  note         = {\url{https://ccam-sunrise-project.eu/}}
}

@misc{synergies,
  author       = {{SYNERGIES Project}},
  title        = {{SYNERGIES}: Real and synthetic scenarios generated for the development, training, virtual testing and validation of {CCAM} systems},
  howpublished = {Horizon Europe project, Grant Agreement No.~101146542},
  note         = {\url{https://synergies-ccam.eu/}}
}

\end{document}